\documentclass[11pt,a4paper]{article}
\usepackage{times,latexsym}
\usepackage{url}
\usepackage[T1]{fontenc}
\usepackage{graphicx}
\usepackage{algorithm}
\usepackage[noend]{algpseudocode}
\usepackage{amsmath}
\usepackage{times}
\usepackage{latexsym}
\usepackage{verbatim}
\usepackage{tabularx}

\newcommand{\hlc}[2][yellow]{{%
    \colorlet{foo}{#1}%
    \sethlcolor{foo}\hl{#2}}%
}
\usepackage{xcolor}
\usepackage{soul}

\usepackage{array}

\newcolumntype{M}[1]{>{\centering\let\newline\\\arraybackslash\hspace{0pt}}m{#1}}
\newcolumntype{L}[1]{>{\centering\let\newline\\\arraybackslash\hspace{0pt}}m{#1}}
\newcolumntype{C}[1]{>{\centering\let\newline\\\arraybackslash\hspace{0pt}}m{#1}}
\newcolumntype{R}[1]{>{\centering\let\newline\\\arraybackslash\hspace{0pt}}m{#1}}
\newcolumntype{P}[1]{>{\centering\let\newline\\\arraybackslash\hspace{0pt}}m{#1}}

   \usepackage[acceptedWithA]{tacl2018v2}
\usepackage{xspace,mfirstuc,tabulary}

\newif\iftaclinstructions
\taclinstructionstrue 
\iftaclinstructions
\renewcommand{\confidential}{}
\renewcommand{\anonsubtext}{(No author info supplied here, for consistency with TACL-submission anonymization requirements)}
\newcommand{\instr}
\fi

\iftaclpubformat 

\else

\fi

\definecolor{KDpurple}{rgb}{0.6,0.18,0.64}

\newcommand*{\affaddr}[1]{#1} 
\newcommand*{\affmark}[1][*]{\textsuperscript{#1}}
\newcommand*{\email}[1]{\texttt{#1}}

\title{Saying No is An Art: Contextualized Fallback Responses for Unanswerable Dialogue Queries}

\author{%
Ashish Shrivastava\affmark[1], Kaustubh D. Dhole\affmark[1], Abhinav Bhatt\affmark[2], Sharvani Raghunath\affmark[1]\\
\affaddr{\affmark[1]Amelia Science, IPsoft R\&D} \hspace{2cm} \affaddr{\affmark[2]Universität des Saarlandes}\\
\email{\affmark[1]\{firstname.lastname\}@ipsoft.com}\\ \email{\affmark[2]abbh00001@stud.uni-saarland.de}
}

\date{12/03/2020}

\begin{document}
\maketitle
\begin{abstract}
Despite end-to-end neural systems making significant progress in the last decade for task-oriented as well as chit-chat based dialogue systems, most dialogue systems rely on hybrid approaches which use a combination of rule-based, retrieval and generative approaches for generating a set of ranked responses. Such dialogue systems need to rely on a fallback mechanism to respond to out-of-domain or novel user queries which are not answerable within the scope of the dialog system. While, dialog systems today rely on static and unnatural responses like ``I don’t know the answer to that question'' or ``I’m not sure about that", we design a neural approach which generates responses which are contextually aware with the user query as well as say no to the user. Such customized responses provide paraphrasing ability and contextualization as well as improve the interaction with the user and reduce dialogue monotonicity. Our simple approach makes use of rules over dependency parses and a text-to-text transformer fine-tuned on synthetic data of question-response pairs generating highly relevant, grammatical as well as diverse questions. We perform automatic and manual evaluations to demonstrate the efficacy of the system.
\end{abstract}

\section{Introduction}

In order to cater to the diversity of questions spanning across various domains, dialogue systems generally follow a hybrid architecture wherein an ensemble of individual response subsystems~\cite{kuratovdream,harrison2020athena} are employed from which an appropriate response is presented to the user~\cite{serban2017deep, finch2020emora, paranjape2020neural}. However, it is common for dialogue systems to encounter queries which are not within their scope of knowledge. While increasing the number of such subsystems would be a good strategy to increase coverage, it can be a never ending process and a default fallback strategy would always be needed. Besides, domain specific dialog systems, especially those deployed in professional settings generally prefer restricting themselves to a fixed set of domains, and purposely refrain from responding to out-of-domain and random or toxic user queries. 

\begin{figure}
\includegraphics[width=\linewidth]{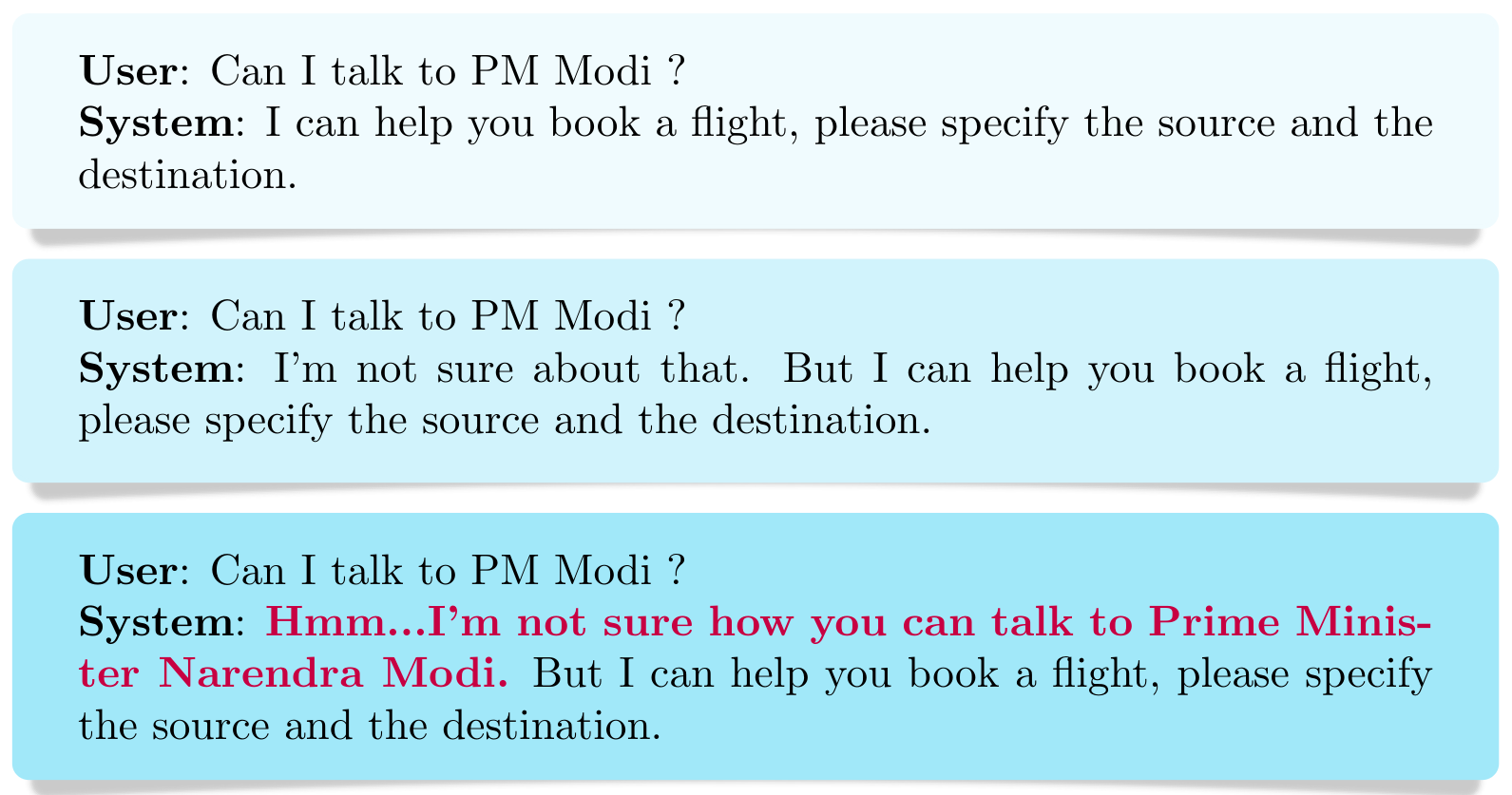}
\caption{Comparison of responses of three flight booking dialog systems: The first one does not handle unknown responses. The second one has a default fall-back response. The third one has a fall-back response which is contextualized with the user query.}
\label{fig:chat_comparison}
\end{figure}

One approach to acknowledge such queries is to have a fallback mechanism with responses like ``I don't know the answer to this question" or ``I'm not sure how to answer that." However, such responses are static and unengaging and give an impression that the user's query has gone unacknowledged or is not understood by the system as shown in Figure~\ref{fig:chat_comparison} above.

\citet{yu-etal-2016-wizard} have shown that static and pre-defined responses lead to lower levels of user engagement and decrease users' interest in interacting with the system. 

Our fallback approach attempts to address these limitations by generating ``don't-know'' responses which are engaging and contextually closer with the user query. 1) Since there are no publicly available datasets to generate such contextualised responses, we synthetically generate (query, fallback response) pairs using a set of highly accurate handcrafted dependency patterns. 2) We then train a sequence-to-sequence model over synthetic and natural paraphrases of these queries. 3) Finally, we measure the grammaticality and relevance of our models using a crowd-sourced setting to assess the generation capability. We publicly release the code and training data used in our experiments.~\footnote{\href{https://github.com/kaustubhdhole/natural-dont-know}{github.com/kaustubhdhole/natural-dont-know}}


\begin{figure*}
\includegraphics[width=\textwidth]{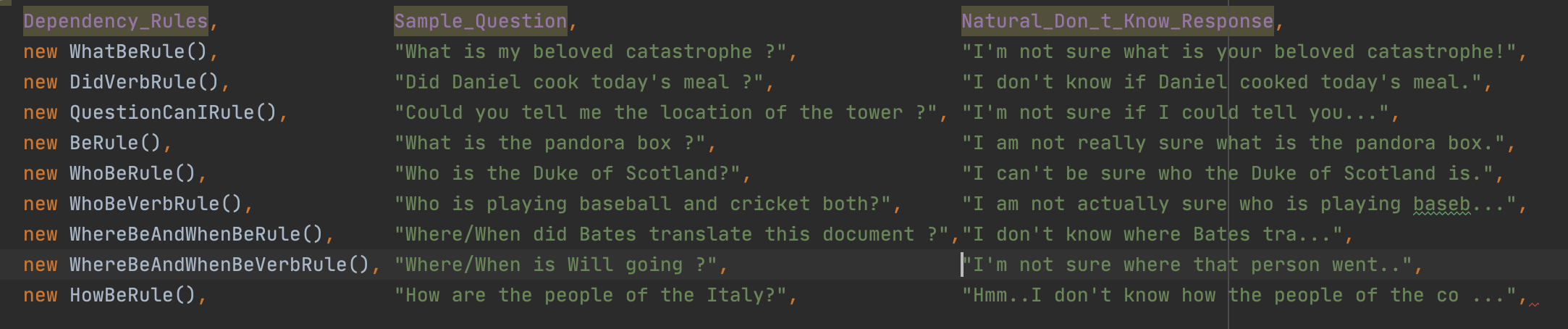}
\caption{Dependency Rules with the class of questions they cater too and their corresponding responses. In the second last sentence, the named entity "Will" is randomly replaced by "that person".}
\label{fig:dep_rules}
\end{figure*}

\section{Related Work}
Improving the coverage to address out-of-domain queries is not a new problem in designing dialog systems. The most popular approach has been via presenting the user with chit-chat responses. Other systems such as Blender~\cite{roller2020recipes} and Meena~\cite{adiwardana2020humanlike} promise to be successful for open-domain settings.~\citet{paranjape2020neural} finetune a GPT-2 model~\cite{radford2019language} on the EmpatheticDialogues dataset~\cite{rashkin2019empathetic} to generate social talk responses. While this might seem fitting for chit-chat and social talk dialog systems, domain-specific scenarios often dealing with professional settings would refrain from performing friendly or social talk especially avoiding the possibility of the randomness of generative models. Also, multiple subsystem architectures always have the possibility of cascading errors and profane or toxic queries. Hence systems always have a foolproof mechanism in the form of static templates to reply from.~\citet{liang2020gunrock} uses an interesting approach for error handling by mapping dialog acts and intents to templates. Besides, like ~\citet{finch2020emora} it is always safer to generate fallback responses on encountering queries which might be toxic, biased or profane. \footnote{Handling programming exceptions and code failures also necessitates a simple fallback approach.}

Another line of work attempts to handle user queries which are ambiguous by asking back clarification questions~\cite{dhole2020resolving, zamani2020generating, yu-etal-2020-interactive}. While this increases user interaction and coverage to an appreciable extent, it does not eliminate the requirement of a failsafe fallback responder. 
This paper's contribution is to address this requirement with an enhanced version of a fallback response generator.

\section{Methods}
We describe two approaches to generate such contextual don't-know responses.

\subsection{The Dependency Based Approach (DBA)}

Inspired by previous approaches which use parse structures to generate questions~\cite{heilman2009question, mazidi-tarau-2016-infusing, dhole-manning-2020-syn}, we create a rule-based generator by handcrafting dependency templates to cater to a wide variety of question patterns as shown in Figure~\ref{fig:dep_rules}. We perform extensive manual testing to improve the generations from these rules and increase overall coverage. The purpose of these rules is two-fold: i) To create a high-precision fall-back response generator as a baseline and ii) to help create (query, don't-know-response) pairs which could be paired with natural paraphrases to serve as seed training data for other deep learning architectures.

To build this baseline generator, we utilize 9 dependency templates in the style of SynQG~\cite{dhole-manning-2020-syn}. We utilize the dependency parser from~\citet{andor2016globally} to get the Universal Dependencies~\cite{nivre2016universal, nivre2017universal} of the user query. We then convert it to a don't-know-response by re-arranging nodes to a matched template. We further change pronouns, incorporate named entity information, and add rules to handle modals and auxiliaries. Finally, we also add rules for flipping pronouns to convert an agent targeted question to a user targeted response by interchanging pronouns and their supporting verbs. E.g. You to I and vice-versa.

We incorporate a bit of paraphrasing by randomizing various prefixes like ``I'm not sure whether'', ``I don't know if'', etc. and randomly using named entities. We describe the high-level algorithm below and in Algorithm~\ref{alg:the_alg}.

\begin{gather*}
prefix = pickRandom(prefixPool) \\
response = DBR(Question)\\
suffix = pickRandom(suffixPool)\\
fallbackResponse = Concat(prefix,\\ response, suffix)
\end{gather*}

\begin{algorithm}[!hbpt]
\caption{Dependency Based Response (DBR)}
  \label{alg:the_alg}
\begin{algorithmic}
\State $nodes \gets \emph{dependencyParse}(Question)$
\For {each template in templatePool}
\If {(template condition matched)}
\State Populate template using \emph{nodes}
\State Handle modals \& auxilliaries
\State Flip pronoun
\State Randomly substitute Named Entity
\EndIf
\EndFor
\If{no template condition matched}
\State $return$ \emph{pickRandom}(defaultResponse pool)
\EndIf
\State $return$ filled template response

\end{algorithmic}
\end{algorithm}

\subsection{Sequence-to-Sequence Approach}
Owing to the expected low coverage and scalability of the rule-based approach, we resort to take advantage of pre-trained neural architectures to attempt to create a sequence-to-sequence fallback responder. To incorporate noise and avoid the model to over-fit on the handcrafted transformations, we do not train the model directly on (query, don't-know-response) pairs generated from the previous section. From all possible questions of the Quora Questions Pairs dataset (QQP)~\footnote{\href{https://www.quora.com/q/quoradata/First-Quora-Dataset-Release-Question-Pairs}{Quora Question Pairs Dataset}}, we first filter all the questions which generate a reply from the dependency based rules. Then we pair these dont-know-responses with the paraphrases of the input questions rather than the input questions themselves.~\footnote{Those question pairs which have the label "1" or are similar are used as paraphrases.} Primarily attempting to avoid over-fitting on the dependency patterns, this also helps generate dont-know-responses which are paraphrastic in nature.

\begin{table}[ht!]
\small
\centering
\begin{tabularx}{0.785\columnwidth}{|l*{3}{c}|l*{3}{c}}
\hline
\hspace{0.25cm}\textbf{Metrics} &
\hspace{0.25cm}\textbf{DBA} &
\hspace{0.25cm}\textbf{Seq-To-Seq} &\\
\hline
\hspace{0.35cm}\%GC &
\hspace{0.2cm}81.6  &
\hspace{0.2cm}87.2 & \\
\hspace{0.4cm}ARS &
\hspace{0.2cm}3.97 &
\hspace{0.2cm}3.66 & \\
\hline
\end{tabularx}
\caption{Human evaluation between the two approaches. \%GC= \% of Grammatically correct responses, ARS=Average Relevance Score. 
  }
\label{human_evaluation-table}
\end{table}

After incorporating paraphrases from QQP, we are able to build a dataset of 100k pairs, which we call the "I Dont Know Dataset" (IDKD). After witnessing the success of text-to-text transformers, we use the pre-trained T5 transformer~\cite{raffel2020exploring} as our sequence-to-sequence model. We divide IDKD into a train and validation split of 80:20. We use the Transformers code from HuggingFace~\cite{wolf-etal-2020-transformers} to fine-tune a T5-base model over IDKD for 2 epochs.~\footnote{Increasing the epochs to 3 and 4 tended to overfit and not generate paraphrases.}

\begin{table*}[ht!]
\small
\centering
\begin{tabular}{|M{4.0cm} L{1.6cm}| C{4.5cm} | R{3.3cm}|}
\hline
\textbf{Metrics} &
\textbf{Question} &
\textbf{Dependency Based Approach} &
\textbf{Seq2Seq Approach}\\
\hline
Coverage &
-- &
\hspace{.2cm}54.56\% &
100\%\\
Average Sentence Length &
9.41  &
11.83 & 
11.78\\
Sentence Length Variance &
11.74 &
11.60 & 
10.66\\
Average \#Novel Words &
-- &
0.91 &
2.57\\

\hline
\end{tabular}
\caption{System evaluation between Dependency Based Approach and Sequence to Sequence Approach.
  }
\label{system_evaluation-table}
\end{table*}

\begin{table*}[ht!]
\small
\centering
\begin{tabular}{|l|}
\hline
\\
Question: What is the quickest way to increase Instagram followers?\\
DBA: I am not sure what the quickest way to increase Instagram followers is.\\
Seq2Seq: I cannot be sure \hlc[cyan!40]{how a user can} increase Instagram followers.\\
\\
Question: How can I join MIT?\\
DBA: I am not sure how you can join MIT.\\
Seq2Seq: I am not sure what \hlc[cyan!40]{the eligibility criteria} for MIT are.\\
\\
Question: How can I drive more traffic to my website?\\
DBA: I'm not really sure how you can drive more traffic.\\
Seq2Seq: I am not sure \hlc[cyan!40]{what the best way to get free traffic} to your website is.\\
\\
Question: Where can I find a collection of poetry by William Shakespeare?\\
DBA: I’m not really sure where you can find a collection of poetry by William Shakespeare.\\
Seq2Seq: I’m not actually sure where you can get a collection of \hlc[cyan!40]{his poems}.\\
\\
Question: What is the effect of demonetization of 500 and 1000 rupees notes?\\
DBA: I cannot be sure what the effect of demonetization of 500 and 1000 rupees notes is.\\
Seq2Seq: I'm not really sure \hlc[cyan!40]{what my views on PM Modi's decision on discontinuing} 500 \& 1000 rs notes are.\\ \\
\hline
\end{tabular}
\caption{Natural don't know responses generated by both the approaches. Highlights in blue depict words, phrases or events not mentioned by the user.}
\label{seq2seq example-table}
\end{table*}
   
\section{Results}
Most prior generated systems are evaluated on a range of automatic metrics like BLEU and ROGUE~\cite{papineni2002bleu} used in the machine translation literature. However, owing to the drawbacks of these metrics, we perform human evaluation of the generated responses using two metrics - namely "relevance" and "grammaticality" as defined in ~\citet{dhole-manning-2020-syn}. We evaluate the performance of both the approaches in a crowd-sourced setting by requesting English-schooled individuals to rate.~\footnote{Our pool of English-schooled raters consisted of engineering graduates having attained high school and engineering education with English as a medium of instruction.} Raters were asked to evaluate grammaticality in a binary setting (grammatical/ungrammatical) and relevance on a Likert scale (1 to 5).

Our human evaluations are shown in Table-\ref{human_evaluation-table}. T5 responses tend to be more grammatical than their dependency counterparts by a large margin of 6\%. Relevance scores drop slightly from 3.97 to 3.66. This can be largely attributed to the model's paraphrastic ability of describing words  and connected events outside the knowledge of the user's query. Eg. in the second query in Table~\ref{seq2seq example-table}, if the string "MIT" were something other than an institution, the dependency based approach would seem safer than the seq2seq approach.  

In addition, T5 responses on an average generate at least double the number of novel words than their dependency counterparts as shown in Table~\ref{system_evaluation-table}. Sentence length mostly remains unaffected across the two models. Undoubtedly, the rule-based model despite being highly relevant is only able to reply to 54.5\% of random QQP queries.

The T5 model helped to not only add paraphrastic variations but also scale to user queries outside of the scope of the dependency templates. More importantly, without losing the original ability of saying no, the model was able to generate more natural sounding dont-know-reponses by utilizing it's inherent world-knowledge acquired during pre-training. Table~\ref{seq2seq example-table} shows some interesting examples. The highlighted phrases in blue show the benefits of the model's pre-training ability.

\section{Conclusion and Future work}
We describe two simple approaches which enhance user interaction to cater to the necessities of real-life dialogue systems which are generally a tapestry of multiple solitary subsystems. In order to avoid cascading errors from such systems, as well as refrain from answering out-of-domain and toxic queries it is but natural to have a fallback approach to say no. We argue that such a fallback approach could be contextualised to generate engaging responses by having multiple ways of saying no rather than a one common string for all approach. The appeal of our approach is the ease with which it can rightly fit within any larger dialog design framework.\\
\indent Of course, this is not to deny that as we give more paraphrasing power to the fallback system, it would tend to retract from succinctly replying with a no - as is evident from the drop in the relevance scores. Nevertheless, we still believe that both our fallback approaches could serve as effective baselines for future work. 

\section{Acknowledgments}
This work was carried out at Amelia R\&D, Bangalore, India 560066.
\bibliography{tacl2018}
\bibliographystyle{acl_natbib}

\end{document}